\documentclass[letterpaper, journal, twoside]{IEEEtran}
\usepackage[utf8]{inputenc}
\usepackage[T1]{fontenc}
\IEEEoverridecommandlockouts   
\usepackage{cite}
\usepackage{amsmath,amssymb,amsfonts}
\usepackage{algorithmic}
\usepackage{array}
\usepackage{graphicx}
\usepackage{textcomp}
\usepackage[table]{xcolor}
\usepackage{booktabs}
\usepackage{makecell}
\usepackage{tikz}
\usepackage{stfloats}          
\usepackage{balance}
\usepackage[group-separator={,}, per-mode=symbol]{siunitx}
\DeclareMathOperator*{\argmin}{argmin}
\newcommand{\ieeepar}[1]{\vspace{0.5em} \noindent\textbf{#1}}

\makeatletter
\let\NAT@parse\undefined
\makeatother
\usepackage{hyperref}
\usepackage{cleveref}
\hypersetup{
    colorlinks=true,
    linkcolor=blue,
    filecolor=magenta,
    urlcolor=cyan,
    pdfpagemode=FullScreen,
}
\title{UNRIO: Uncertainty-Aware Velocity Learning \\for Radar-Inertial Odometry}

\author{Jui-Te Huang, Tianshu Huang, Anthony Rowe, and Michael Kaess
\thanks{J. Huang, T. Huang, A. Rowe and M. Kaess are with the Carnegie Mellon University, Pittsburgh, PA 15213, USA \texttt{\{juiteh, tianshu2, agr, kaess\}@andrew.cmu.edu}}%
}

\begin{document}


\maketitle


\newcommand{\vc}[1]{\boldsymbol{#1}}
\newcommand{\adj}[1]{\frac{d J}{d #1}}
\newcommand{\chain}[2]{\adj{#2} = \adj{#1}\frac{d #1}{d #2}}

\newcommand{\Ac}{\mathcal{A}}
\newcommand{\Bc}{\mathcal{B}}
\newcommand{\Cc}{\mathcal{C}}
\newcommand{\Dc}{\mathcal{D}}
\newcommand{\Ec}{\mathcal{E}}
\newcommand{\Fc}{\mathcal{F}}
\newcommand{\Gc}{\mathcal{G}}
\newcommand{\Hc}{\mathcal{H}}
\newcommand{\Ic}{\mathcal{I}}
\newcommand{\Jc}{\mathcal{J}}
\newcommand{\Kc}{\mathcal{K}}
\newcommand{\Lc}{\mathcal{L}}
\newcommand{\Mc}{\mathcal{M}}
\newcommand{\Nc}{\mathcal{N}}
\newcommand{\Oc}{\mathcal{O}}
\newcommand{\Pc}{\mathcal{P}}
\newcommand{\Qc}{\mathcal{Q}}
\newcommand{\Rc}{\mathcal{R}}
\newcommand{\Sc}{\mathcal{S}}
\newcommand{\Tc}{\mathcal{T}}
\newcommand{\Uc}{\mathcal{U}}
\newcommand{\Vc}{\mathcal{V}}
\newcommand{\Wc}{\mathcal{W}}
\newcommand{\Xc}{\mathcal{X}}
\newcommand{\Yc}{\mathcal{Y}}
\newcommand{\Zc}{\mathcal{Z}}

\newcommand{\Ab}{\mathbb{A}}
\newcommand{\Bb}{\mathbb{B}}
\newcommand{\Cb}{\mathbb{C}}
\newcommand{\Db}{\mathbb{D}}
\newcommand{\Eb}{\mathbb{E}}
\newcommand{\Fb}{\mathbb{F}}
\newcommand{\Gb}{\mathbb{G}}
\newcommand{\Hb}{\mathbb{H}}
\newcommand{\Ib}{\mathbb{I}}
\newcommand{\Jb}{\mathbb{J}}
\newcommand{\Kb}{\mathbb{K}}
\newcommand{\Lb}{\mathbb{L}}
\newcommand{\Mb}{\mathbb{M}}
\newcommand{\Nb}{\mathbb{N}}
\newcommand{\Ob}{\mathbb{O}}
\newcommand{\Pb}{\mathbb{P}}
\newcommand{\Qb}{\mathbb{Q}}
\newcommand{\Rb}{\mathbb{R}}
\newcommand{\Sb}{\mathbb{S}}
\newcommand{\Tb}{\mathbb{T}}
\newcommand{\Ub}{\mathbb{U}}
\newcommand{\Vb}{\mathbb{V}}
\newcommand{\Wb}{\mathbb{W}}
\newcommand{\Xb}{\mathbb{X}}
\newcommand{\Yb}{\mathbb{Y}}
\newcommand{\Zb}{\mathbb{Z}}

\newcommand{\av}{\mathbf{a}}
\newcommand{\bv}{\mathbf{b}}
\newcommand{\cv}{\mathbf{c}}
\newcommand{\dv}{\mathbf{d}}
\newcommand{\ev}{\mathbf{e}}
\newcommand{\fv}{\mathbf{f}}
\newcommand{\gv}{\mathbf{g}}
\newcommand{\hv}{\mathbf{h}}
\newcommand{\iv}{\mathbf{i}}
\newcommand{\jv}{\mathbf{j}}
\newcommand{\kv}{\mathbf{k}}
\newcommand{\lv}{\mathbf{l}}
\newcommand{\mv}{\mathbf{m}}
\newcommand{\nv}{\mathbf{n}}
\newcommand{\ov}{\mathbf{o}}
\newcommand{\pv}{\mathbf{p}}
\newcommand{\qv}{\mathbf{q}}
\newcommand{\rv}{\mathbf{r}}
\newcommand{\sv}{\mathbf{s}}
\newcommand{\tv}{\mathbf{t}}
\newcommand{\uv}{\mathbf{u}}
\newcommand{\vv}{\mathbf{v}}
\newcommand{\wv}{\mathbf{w}}
\newcommand{\xv}{\mathbf{x}}
\newcommand{\yv}{\mathbf{y}}
\newcommand{\zv}{\mathbf{z}}

\newcommand{\Av}{\mathbf{A}}
\newcommand{\Bv}{\mathbf{B}}
\newcommand{\Cv}{\mathbf{C}}
\newcommand{\Dv}{\mathbf{D}}
\newcommand{\Ev}{\mathbf{E}}
\newcommand{\Fv}{\mathbf{F}}
\newcommand{\Gv}{\mathbf{G}}
\newcommand{\Hv}{\mathbf{H}}
\newcommand{\Iv}{\mathbf{I}}
\newcommand{\Jv}{\mathbf{J}}
\newcommand{\Kv}{\mathbf{K}}
\newcommand{\Lv}{\mathbf{L}}
\newcommand{\Mv}{\mathbf{M}}
\newcommand{\Nv}{\mathbf{N}}
\newcommand{\Ov}{\mathbf{O}}
\newcommand{\Pv}{\mathbf{P}}
\newcommand{\Qv}{\mathbf{Q}}
\newcommand{\Rv}{\mathbf{R}}
\newcommand{\Sv}{\mathbf{S}}
\newcommand{\Tv}{\mathbf{T}}
\newcommand{\Uv}{\mathbf{U}}
\newcommand{\Vv}{\mathbf{V}}
\newcommand{\Wv}{\mathbf{W}}
\newcommand{\Xv}{\mathbf{X}}
\newcommand{\Yv}{\mathbf{Y}}
\newcommand{\Zv}{\mathbf{Z}}

\newcommand{\alphav     }{\boldsymbol \alpha     }
\newcommand{\betav      }{\boldsymbol \beta      }
\newcommand{\gammav     }{\boldsymbol \gamma     }
\newcommand{\deltav     }{\boldsymbol \delta     }
\newcommand{\epsilonv   }{\boldsymbol \epsilon   }
\newcommand{\varepsilonv}{\boldsymbol \varepsilon}
\newcommand{\zetav      }{\boldsymbol \zeta      }
\newcommand{\etav       }{\boldsymbol \eta       }
\newcommand{\thetav     }{\boldsymbol \theta     }
\newcommand{\varthetav  }{\boldsymbol \vartheta  }
\newcommand{\iotav      }{\boldsymbol \iota      }
\newcommand{\kappav     }{\boldsymbol \kappa     }
\newcommand{\varkappav  }{\boldsymbol \varkappa  }
\newcommand{\lambdav    }{\boldsymbol \lambda    }
\newcommand{\muv        }{\boldsymbol \mu        }
\newcommand{\nuv        }{\boldsymbol \nu        }
\newcommand{\xiv        }{\boldsymbol \xi        }
\newcommand{\omicronv   }{\boldsymbol \omicron   }
\newcommand{\piv        }{\boldsymbol \pi        }
\newcommand{\varpiv     }{\boldsymbol \varpi     }
\newcommand{\rhov       }{\boldsymbol \rho       }
\newcommand{\varrhov    }{\boldsymbol \varrho    }
\newcommand{\sigmav     }{\boldsymbol \sigma     }
\newcommand{\varsigmav  }{\boldsymbol \varsigma  }
\newcommand{\tauv       }{\boldsymbol \tau       }
\newcommand{\upsilonv   }{\boldsymbol \upsilon   }
\newcommand{\phiv       }{\boldsymbol \phi       }
\newcommand{\varphiv    }{\boldsymbol \varphi    }
\newcommand{\chiv       }{\boldsymbol \chi       }
\newcommand{\psiv       }{\boldsymbol \psi       }
\newcommand{\omegav     }{\boldsymbol \omega     }

\newcommand{\Gammav     }{\boldsymbol \Gamma     }
\newcommand{\Deltav     }{\boldsymbol \Delta     }
\newcommand{\Thetav     }{\boldsymbol \Theta     }
\newcommand{\Lambdav    }{\boldsymbol \Lambda    }
\newcommand{\Xiv        }{\boldsymbol \Xi        }
\newcommand{\Piv        }{\boldsymbol \Pi        }
\newcommand{\Sigmav     }{\boldsymbol \Sigma     }
\newcommand{\Upsilonv   }{\boldsymbol \Upsilon   }
\newcommand{\Phiv       }{\boldsymbol \Phi       }
\newcommand{\Psiv       }{\boldsymbol \Psi       }
\newcommand{\Omegav     }{\boldsymbol \Omega     }

\begin{abstract}

mmWave radars are robust to darkness and occlusions such as dust and smoke, and can directly constrain ego-velocity from a single frame via Doppler measurements, making them attractive sensors for odometry in visually denied conditions. However, almost all existing radar-inertial odometry systems rely on lossy radar point clouds that are highly sparse, generally concentrated in a narrow angular band, and aliased at high speeds. We propose to instead estimate ego-velocity directly from unfiltered mmWave I/Q signals. Taking advantage of a foundation model for 4D radar spectrum, we develop a system that integrates uncertainty-aware velocity predictions with IMU measurements using an uncertainty-weighted sliding-window pose graph to accurately compute odometry even when provided radar data with aliasing or an unfavorable field of view. Evaluated on public benchmark datasets, our system, UNRIO, attains the lowest relative pose error on the majority of sequences across held-out environments, despite differing chirp configurations, motion patterns, and platforms — with its strongest gains coming where point clouds fail most: lateral motion in IQ1M and heavily aliased Doppler in ColoRadar.


\end{abstract}


\section{Introduction}

\IEEEPARstart{M}{illimeter-wave} (mmWave) radar is an attractive sensor for robot state estimation. As an active sensor, radars are unaffected by darkness; in addition, its millimeter-scale wavelength penetrates dust, smoke, and fog, allowing it to remain operational even where LiDAR and RGB cameras fail. Modern single-chip radars are also inexpensive, compact, and low-power, allowing them to be readily deployed on handheld and aerial platforms~\cite{iovescu2017fundamentals}.

Furthermore, in contrast to other sensors used for state estimation, radar is uniquely able to measure motion directly. The radial velocity of each reflector can be measured via its Doppler shift on the reflected signal, allowing a single frame to constrain ego-velocity without the cross-frame feature association or scan alignment that cameras and LiDARs require. This has driven a large body of work on radar-inertial state estimation: filtering and windowed optimization that fuse a single-scan velocity estimate with IMU preintegration~\cite{DoerMFI2020, doer2021x, michalczyk2023multi, kim2025ekf, huang2024multi}, and, as commercial radars gained angular resolution, full SLAM systems built on radar scan matching~\cite{zhang20234dradarslam, zhuang20234d, herraez2025rai}.


\begin{figure}[t]
    \centering
    \includegraphics[width=\linewidth]{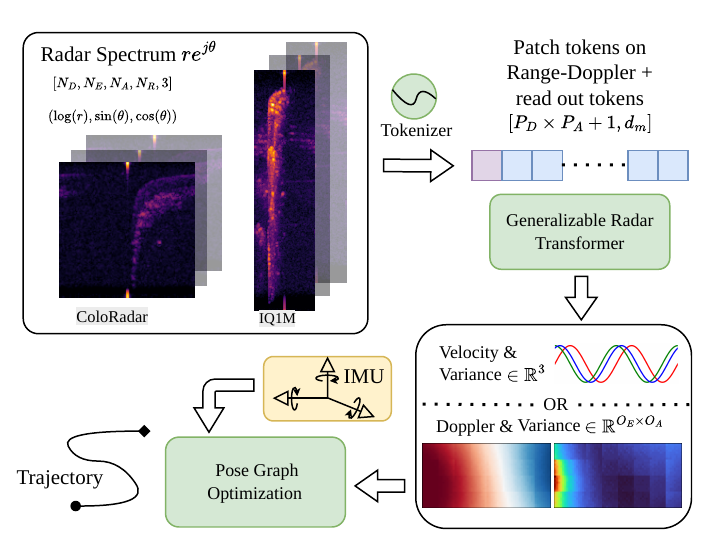}
    \vspace*{-6mm}
    \caption{\textbf{Our radar-inertial odometry system uses raw mmWave radar spectrum as input.} Our system predicts velocity and its uncertainty from the radar spectrum using a Transformer neural network trained on a large dataset with diverse motion patterns. ColoRadar longboard ride at 4.6 m/s, past the waveform's $\pm2.50$ m/s unambiguous interval: the static-scene return folds across the Doppler axis.}
    \label{fig:teaser}
\end{figure}

These methods apply classical geometric estimators to 4D radar point clouds which are derived from raw radar signals using Constant False Alarm Rate (CFAR) peak detectors combined with Angle-of-Arrival estimation techniques~\cite{iovescu2017fundamentals}. However, radar point clouds are extremely lossy (Fig.~\ref{fig:rd_cfar}). Detections are sparse --- typically on the order of $\approx$100 points per frame --- and concentrated in a particular angular band where the antenna's gain is strong enough to generate clear peaks. Finally, speeds beyond a maximum \textit{unambiguous Doppler interval} determined by the radar modulation alias and ``wrap around,'' leading to a biased velocity estimate.


In contrast, Radar spectrum retains all information measured by the radar sensor. Crucially, in radar spectrum, ego-velocity is a property of the global structure across all bins --- not just a handful of peaks, so remains observable even when no single return is strong enough to be detected. Geometric context across the frame also carries evidence which can disambiguate Doppler velocity aliasing which cannot be done with just a single peak. While this is difficult to extract analytically, recent work also supports that this information is recoverable in practice: networks trained directly on the radar spectrum outperform point-cloud pipelines across perception tasks~\cite{zhang2021raddet, giroux2023t, decourt2022darod, prabhakara2023high, li2023azimuth, lai2024enabling}, and foundation-model-style pretraining shows that performance improves with data size~\cite{huang2025towards}.


Mirroring recent advances in visual odometry, where learned features  and confidence weights are fed into a classical optimization back end~\cite{teed2021droid, wang2024dust3r, qiu2025mac}, we propose to combine a learned, uncertainty-aware radar spectrum-based model with a pose graph optimization algorithm to improve robustness and accuracy while keeping the geometry explicit. Our contributions are as follows:

\begin{itemize}
    \item We develop a learning-based method that estimates ego-velocity directly from raw radar spectrum data, supporting two output modes: a single linear velocity estimate and a per-angle-bin Doppler velocity map (Sec.~\ref{sec:nn_arch}). 
    
    \item We extend the model to predict uncertainty alongside the velocity estimate, which is subsequently propagated into our radar-inertial odometry framework (Sec.~\ref{sec:nn_train}). 

    \item We train our model on a large-scale raw radar spectrum dataset and introduce UNRIO, a radar-inertial odometry algorithm. We evaluate its state estimation performance on held-out environments with diverse forward and lateral motion patterns, and show that it transfers to a different chirp configuration and platform after fine-tuning on a separate dataset (Sec.~\ref{sec:exp}).


\end{itemize}

\ifdefined\rdw\else\newlength{\rdw}\fi
\ifdefined\rdh\else\newlength{\rdh}\fi

\begin{figure}[t]
  \centering
  \setlength{\rdw}{0.85\linewidth}%
  \setlength{\rdh}{0.25\rdw}
  \begin{tikzpicture}[
      img/.style={inner sep=0pt, outer sep=0pt},
      tk/.style={font=\scriptsize, inner sep=2pt, overlay},
      ax/.style={font=\footnotesize, inner sep=2pt, overlay},
    ]
    \node[img, anchor=north west] (a) at (0,0)
      {\includegraphics[width=\rdw]{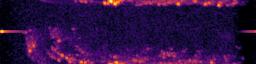}};
    \node[img, anchor=north west] (b) at (0,-\rdh-4pt)
      {\includegraphics[width=\rdw]{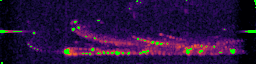}};

    \draw[->, white, line width=2pt]
         ([xshift=0.45\rdw, yshift=-0.4\rdh]a.north west)
      -- ([xshift=0.45\rdw, yshift=-0.17\rdh]a.north west);
    \draw[->, white, line width=2pt]
         ([xshift=0.45\rdw, yshift=-0.68\rdh]a.north west)
      -- ([xshift=0.45\rdw, yshift=-0.91\rdh]a.north west);
             
    \node[tk, anchor=north west] at (a.north east) {$+1.2$};
    \node[tk, anchor=west]       at (a.east)       {$0$};
    \node[tk, anchor=south west] at (a.south east) {$-1.2$};
    \node[tk, anchor=north west] at (b.north east) {$+1.2$};
    \node[tk, anchor=west]       at (b.east)       {$0$};
    \node[tk, anchor=south west] at (b.south east) {$-1.2$};

    \node[tk, anchor=north west] at (b.south west) {$0$};
    \node[tk, anchor=north east] at (b.south east) {$11.2$};

    \node[ax, rotate=90, anchor=south] at (-0.5em,-\rdh) {Doppler (m/s)};
    \node[ax, anchor=north] at (b.south) {Range (m)};
  \end{tikzpicture}

  \caption{%
    Two ways radar point clouds lose information. Top: at $1.45$~m/s, the platform outruns the
    $\pm1.22$~m/s unambiguous interval, so the return folds across the
    Doppler axis. Bottom: green marks the $1.8\%$ of bins CFAR keeps,
    which are all the body-velocity optimization sees.%
  }
  \label{fig:rd_cfar}
  \vspace{-5mm}
\end{figure}

\section{Related Work}

\subsection{Radar-Inertial Odometry}

Taking advantage of the active, occlusion-resistant nature of Radar, Doer et al.~\cite{DoerMFI2020, doer2021x} first proposed building radar-inertial odometry systems by estimating body-frame velocity from radar point clouds combined with IMU via EKF, which has been followed by extensions to include online temporal calibration~\cite{kim2025ekf} and an alternative fixed-lag window optimization strategy~\cite{kim2025ekf}. As recent commercial mmWave radars with more antennas and higher angular resolution have become available, other works have also begun to incorporate scan-matching strategies from LiDAR SLAM systems \cite{zhuang20234d, herraez2025rai}.

Recent works have also begun to explore the effects of signal processing on radar odometry~\cite{jiang2025digital} and even learning velocity and uncertainty directly from raw spectrum using neural networks \cite{rai2025uncertainty}. However, despite these developments, no existing work has attempted to leverage full radar spectrum information for an odometry system.

\subsection{Machine Learning for Radar Spectrum}


While most radar odometry systems — and radar perception systems more broadly — rely on point clouds, point cloud extraction discards much of the information contained in raw radar spectrum data. Competing approaches instead propose to directly interpret the cube by training neural networks with various architectures on large volumes of data, and are able to achieve superior resolution and accuracy on tasks ranging from object detection \cite{zhang2021raddet,giroux2023t,decourt2022darod} and super-resolution \cite{prabhakara2023high,huang2025towards} to camera-like semantic segmentation \cite{lai2024enabling,huang2025towards} and multi-view 3D reconstruction \cite{huang2024dart,borts2024radar,li2023azimuth}.

In particular, models trained using foundation model-style pretraining have been shown to be highly scalable with increased training data \cite{huang2025towards}. Increasingly capable edge compute systems have also made deploying ML-based radar perception to real-time systems \cite{dowling2026centralized} much more practical. Together, this raises the possibility of using large-scale datasets to pretrain powerful, general-purpose models for radar perception, which are then distilled and deployed to autonomous robots and vehicles. We explore one downstream application using models built for radar spectrum in this work.

\subsection{Uncertainty Prediction for State Estimation}

Sensor measurements in the real world are rarely uniform in quality. Particularly in state estimation, we care about how much each measurement can be trusted, using its uncertainty to weight that measurement so unreliable ones contribute less to the solution. While this uncertainty is traditionally precalibrated offline or computed analytically on the fly from a measurement model, emerging learning-based methods instead predict a covariance per measurement; this is a natural fit when the measurement is itself deduced by a learned model.

In the recent SLAM literature, this prediction takes three forms: (1) implicitly, by differentiating through the optimizer~\cite{teed2021droid}, where the weights have no target of their own; (2) explicitly as a relative confidence, a unitless weight trained jointly with a geometric prediction~\cite{wang2024dust3r, murai2025mast3r}, which orders measurements but carries no scale; and (3) explicitly as a covariance in the units of the residual it weights~\cite{dexheimer2023learning, qiu2025mac}, which the optimizer can consume as a noise model directly. We follow the metric-scale route, predicting the ego-velocity uncertainty in $(\text{m/s})^2$ so the pose graph can consume it without recalibration.


\section{Methodology}
UNRIO (Fig.~\ref{fig:teaser}) uses a GRT-style \cite{huang2025towards} model to estimate body-frame velocity and its uncertainty from mmWave radar spectrum. These predictions are then fused with IMU measurements in a sliding-window pose graph optimization to recover the odometry trajectory.

\subsection{Radar Signal Processing (RSP)}
\label{sec:signal_processing}

UNRIO uses a typical commercially available mmWave radar that transmits frequency-modulated continuous-wave (FMCW) which emits a sequence of chirps across an antenna array in a multi-input multi-output (MIMO) mode. The raw output of a single frame is a complex IQ tensor of shape $\left(N_c, N_{tx}, N_{rx}, N_r\right)$, where $N_c$ is the number of chirps (slow-time), $N_{tx}$ and $N_{rx}$ are the transmit and receive antenna counts, and $N_r$ is the number of fast-time samples per chirp. 

To process this data, we apply a four-dimensional Fast Fourier Transform (FFT). A range FFT along the fast-time axis and a Doppler FFT along the slow-time axis produce a range-Doppler map. Subsequently, virtual aperture synthesis via elevation and azimuth FFTs produces a 4-D spectral cube  $\mathbf{S} \in \mathbb{C}^{N_D \times N_E \times N_A \times N_R}$, where $N_D$, $N_E$, $N_A$, and $N_R$ denote Doppler, elevation, azimuth, and range bin counts, respectively. We pad the FFT size to $64$ for both angular FFTs to contain more detailed angular information. 

After the FFTs, each complex bin $re^{j\theta}$ is then encoded as a three-channel feature $\left(\log_{10}r,\; sin\theta,\; \cos\theta \right)$, separating log-magnitude from the phase components\footnote{
Providing $\sin$ and $\cos$ components represents phase angle information without any discontinuities; this differs from GRT~\cite{huang2025towards}, which encodes the phase component as an angle in $[-\pi, \pi)$ with a discontinuity at $\pm \pi$.
}, giving the network input $\mathbf{X} \in \mathbb{R}^{N_D \times N_E \times N_A \times N_R \times 3}$.

\subsection{Neural Network Architecture}
\label{sec:nn_arch}
Our velocity and uncertainty prediction network is based on the GRT~\cite{huang2025towards} transformer architecture. We experiment with two types of velocity prediction: directly outputting the linear body-frame velocity, and predicting per-angle-bin Doppler values. To enable uncertainty estimation, we also design the network to have separate decoder heads for prediction and uncertainty outputs.

\ieeepar{Tokenizer.}
The spectral cube $\mathbf{X}$ is processed by a GRT-style spectrum tokenizer to patchify the inputs into tokens. The elevation and azimuth axes are first collapsed into the channel dimension, after which spatial patching is applied over the Doppler and range axes with a linear projection to the model's channel width. Sinusoidal positional encodings are then added, and a learned readout token is appended.

\ieeepar{Encoder.}
A 4-layer multi-head self-attention transformer encoder processes tokens $\mathbf{T}$ into contextual features $\mathbf{E}$.

\ieeepar{Depth and Doppler decoder.}
A 4-layer Transformer decoder uses a learned query grid of the desired output image of size $(O_E, O_A)$, corresponding to the required output angle bins. The queries are then cross-attended to encoded tokens $\Ev$ and produce an output image with 2 channels: a normalized depth map $\hat{\rho}\in\Rb^{O_E\times O_A}$ and a normalized Doppler image $\hat{\dv}\in\Rb^{O_E\times O_A}$ where the maximum range and unambiguous Doppler for the radar correspond to a value of $+ 1$.

\ieeepar{Uncertainty decoder.}
A separate 4-layer Transformer decoder with architecture identical to the prediction decoder produces a per-pixel depth log-variance map $\hat{\boldsymbol{\ell}}_r\in\Rb^{O_e\times O_a}$. Every log-variance the network emits is passed through a scaled logistic that bounds it to $(\ell_{\min},\ell_{\max})=(-12,6)$,
\begin{equation}
  \hat{\ell} = \ell_{\min} + (\ell_{\max}-\ell_{\min})\,
    \mathrm{sigmoid}\!\left(\frac{4\,z}{\ell_{\max}-\ell_{\min}}\right),
  \label{eq:softclamp}
\end{equation}
where $z$ is the raw output and the factor $4/(\ell_{\max}-\ell_{\min})$ makes the slope unity at $z=0$. Unlike a hard clamp, the gradient never vanishes at the bounds, so a saturated pixel can still recover during training.

\ieeepar{Velocity and uncertainty heads.}
Two 3-layer MLPs operate on the encoder output readout token. The velocity head outputs a positive velocity scale and a unit direction. $[\sv, \nv] \in \Rb^4$, which then converted to body-frame velocity as $\hat{\vv}=\sv\nv \in \Rb^3$. The log-variance head outputs per-axis log-variance $\hat{\boldsymbol{\ell}}_v \in \mathbb{R}^3$, also softly clamped.

\subsection{Training Strategy}
\label{sec:nn_train}
Following modern foundation model best-practices, we train the UNRIO backbone in three stages, with each building on the trained model weights from the previous stage.

\ieeepar{Stage 1 --- Depth pretraining.}
To provide strong geometric pre-training before Doppler or velocity supervision is applied, the tokenizer, encoder, and decoder are first trained to predict LiDAR-projected depth using a masked L1 loss and a spatial gradient regularizer
\begin{equation}
  \Lc_\mathrm{depth}
    = \Lc_\mathrm{L1}(\hat{\rho}, \rho^\star, \Vc)
    + \lambda_g \Lc_\mathrm{grad}(\hat{\rho}, \rho^\star, \Vc),
    \label{eq:depth_gt}
\end{equation}
where $\rho^\star$ is the normalized LiDAR depth by maximum radar distance, $\Vc$ is a valid-pixel mask within the maximum radar range, and $\lambda_g{=}0.5$ weights the gradient term. In this stage, we train the network with all available data from the GRT dataset, including three different chirp configurations, depth, and velocity patterns.

\ieeepar{Stage 2 --- Velocity or Doppler prediction.}
\label{sec:vel_pred}
In this stage, we fine-tune on our target indoor environment subset of the GRT dataset. Depending on the experiment, we either trained on velocity or Doppler output heads. For training velocity $\hat{\vv}$, we simply use an L1 loss with ground truth body velocity $\vv^\star$ calculated from LiDAR odometry. For training the Doppler image $\hat{\dv}$, we project the ground truth body velocity to each angle bin and obtain the ground truth Doppler image
\begin{equation}
  \dv^{\star}_{e,a} = -\, \vv^\star \cdot \hat{\uv}_{e,a}  
\end{equation}
Where $\hat{\uv}_{e,a}\in\Rb^3$ be the unit direction of the angle bin $[e,a]$. The uncertainty decoder is still frozen in this stage.

\ieeepar{Stage 3 --- Uncertainty calibration.}
\label{sec:vel_cov}
Finally, we train our model to output prediction uncertainty, with all model components frozen except the uncertainty decoder or uncertainty heads being trained. We follow previous works on metric scale uncertainty prediction for state estimation\cite{qiu2025mac}, which use the negative log-likelihood loss:
\begin{equation}
  \Lc_\mathrm{NLL}
    = \lambda_\mathrm{nll}
      \sum_{i}\frac{(est_i - gt_i)^2}{2\hat{\sigma}^2_{i}}
      + \frac{1}{2}\log \hat{\sigma}^2_{i}
\end{equation}
where $(est,gt)$ can be either velocity $(\hat{\vv},\vv^\star)$ or metric scale depth image $(\hat{\rv}_m,\rv^\star)$, and $\hat{\sigma}^2_{i}$ is correspondingly either the per-axis velocity variance $\exp(\hat{\boldsymbol{\ell}}_v)$ or the per-pixel depth variance $\exp(\hat{\boldsymbol{\ell}}_r)$. Since the backbone weights are frozen and $est$ is detached, the gradients flow exclusively from the NLL loss through the uncertainty model, training it to predict uncertainty that is calibrated to the actual prediction error.

Note that we calibrate uncertainty on depth rather than on Doppler: depth is supervised by dense LiDAR ground truth at every angle bin, whereas the Doppler ground truth (Eq.~\ref{eq:depth_gt}) is a rigid-body projection that carries no per-pixel notion of measurement quality. As such, a per-pixel Doppler variance estimate trained against it would mostly fit the projection residual instead of the sensing reliability.

\subsection{Velocity and Uncertainty Propagate}
While we can directly use the predicted velocity and its uncertainty $(\hat{\vv}, \hat{\boldsymbol{\ell}}_v)$, where $\Sigma_v=\mathrm{diag}(\exp(\hat{\boldsymbol{\ell}}_v))$, for the radar-inertial odometry task, we need to run one additional step to transform the Doppler image $\hat{\dv}_m$ into a velocity $\hat{\vv}_d\in\Rb^3$ and an associated $3\times3$ matrix $\Sigma_d$. The two heads share a backbone and an angle-bin grid, so a bin whose range is predicted confidently is also a bin where the network has localized a strong, well-conditioned target, and its Doppler estimate is likewise more trustworthy. We therefore reuse the calibrated depth log-variance as a per-pixel \emph{confidence weight} for the Doppler fit,
\begin{equation}
  w(e,a) = \exp\bigl(-\hat{\boldsymbol{\ell}}_r(e,a)\bigr),
\end{equation}
and recover the body-frame velocity by solving the weighted least-squares (WLS) problem
\begin{equation}
  \hat{\vv}_d
    = \arg\min_\vv
      \sum_{e,a} w(e,a)\,
      \bigl[\hat{\dv}_m(e,a) - \vv \cdot \hat{\uv}(e,a)\bigr]^2
\end{equation}
where $\hat{\uv}(e,a)\in\Rb^3$ is the unit direction vector at elevation and azimuth angle. Stacking all $N = O_E\times O_A$ direction vectors as rows of $\mathbf{D} \in \mathbb{R}^{N \times 3}$ and letting $\mathbf{W} = \mathrm{diag}(w_1, \ldots, w_N)$, the normal equations give
\begin{equation}
  \hat{\vv}_d = (\mathbf{D}^\top \mathbf{W} \mathbf{D})^{-1}
                     \mathbf{D}^\top \mathbf{W} \hat{\dv}_m
  \label{eq:wls}
\end{equation}
solved via Cholesky decomposition, and the inverse information matrix
\begin{equation}
  \boldsymbol{\Sigma}_d = (\mathbf{D}^\top \mathbf{W} \mathbf{D})^{-1}
  \label{eq:cov}
\end{equation}
serves as the velocity uncertainty passed to the filter.

We stress that $w$ is a heuristic weighting rather than a Doppler noise model, so Eq.~\eqref{eq:cov} is a covariance only up to an unknown global scale. It still gives the filter what it needs: which angle bins to trust within a frame, and the shape of the field of view, so that velocity directions seen by few or unreliable bins get a larger variance.

\begin{table*}[!t]
    \centering
\centering
\small
\caption{APE / RPE RMSE (m) --- IQ1M sequences with forward/lateral motion. Color background indicate \colorbox{yellow!40}{\strut best}, \colorbox{gray!20}{\strut 2nd best} performance.}
\vspace{-2mm}
\resizebox{0.95\textwidth}{!}{%
\begin{tabular}{lrrrrrr|rrrrr}
\toprule
Trace & Dist (m) & \multicolumn{5}{c}{APE} & \multicolumn{5}{c}{RPE} \\
\cmidrule{3-7} \cmidrule{8-12}
 &  & PC~\cite{huang2024multi} & \makecell{EKF\\RIO}~\cite{DoerMFI2020} & GRT~\cite{huang2025towards} & \makecell{UNRIO\\Doppler} & \makecell{UNRIO\\velocity} & PC~\cite{huang2024multi} & \makecell{EKF\\RIO}~\cite{DoerMFI2020} & GRT~\cite{huang2025towards} & \makecell{UNRIO\\Doppler} & \makecell{UNRIO\\velocity} \\
\midrule
cfa.1.fwd & 64 & \cellcolor{gray!20}0.289 & 0.685 & 0.758 & 0.316 & \cellcolor{yellow!40}0.267 & 0.408 & 0.886 & 0.934 & \cellcolor{gray!20}0.212 & \cellcolor{yellow!40}0.097 \\
cfa.3.fwd & 65 & 0.214 & 0.251 & 0.626 & \cellcolor{gray!20}0.211 & \cellcolor{yellow!40}0.100 & 0.444 & 0.543 & 0.611 & \cellcolor{gray!20}0.207 & \cellcolor{yellow!40}0.151 \\
cfa.a.fwd (1) & 112 & 0.281 & 1.626 & 0.622 & \cellcolor{gray!20}0.237 & \cellcolor{yellow!40}0.221 & 0.589 & 0.950 & 0.458 & \cellcolor{gray!20}0.199 & \cellcolor{yellow!40}0.181 \\
cfa.a.fwd (2) & 127 & \cellcolor{yellow!40}0.148 & 5.597 & 0.998 & 0.903 & \cellcolor{gray!20}0.279 & 0.482 & 1.258 & 0.757 & \cellcolor{gray!20}0.460 & \cellcolor{yellow!40}0.196 \\
morrison.1.fwd (1) & 199 & 3.193 & 3.930 & \cellcolor{yellow!40}1.433 & 3.917 & \cellcolor{gray!20}1.680 & 0.449 & 1.074 & 0.450 & \cellcolor{gray!20}0.212 & \cellcolor{yellow!40}0.148 \\
morrison.1.fwd (2) & 55 & 1.340 & \cellcolor{gray!20}0.140 & 0.532 & 0.304 & \cellcolor{yellow!40}0.129 & 1.560 & 0.440 & 0.575 & \cellcolor{gray!20}0.174 & \cellcolor{yellow!40}0.124 \\
morrison.2.fwd (1) & 91 & 0.959 & 1.224 & 1.022 & \cellcolor{gray!20}0.852 & \cellcolor{yellow!40}0.706 & 0.337 & 0.804 & 0.568 & \cellcolor{gray!20}0.273 & \cellcolor{yellow!40}0.209 \\
morrison.2.fwd (2) & 53 & \cellcolor{gray!20}0.130 & 0.445 & 0.501 & \cellcolor{yellow!40}0.114 & 0.276 & 0.450 & 0.611 & 0.885 & \cellcolor{yellow!40}0.125 & \cellcolor{gray!20}0.323 \\
posner.1.fwd & 335 & \cellcolor{yellow!40}1.938 & 9.374 & 3.020 & 4.794 & \cellcolor{gray!20}2.794 & 1.913 & 1.395 & 0.705 & \cellcolor{gray!20}0.281 & \cellcolor{yellow!40}0.223 \\
posner.2.fwd & 223 & \cellcolor{yellow!40}0.142 & 8.192 & 1.087 & 0.947 & \cellcolor{gray!20}0.896 & 0.439 & 1.025 & 0.408 & \cellcolor{gray!20}0.269 & \cellcolor{yellow!40}0.197 \\
posner.3.fwd & 160 & 1.587 & 6.221 & \cellcolor{gray!20}1.394 & 2.459 & \cellcolor{yellow!40}0.913 & 1.678 & 1.487 & 0.417 & \cellcolor{gray!20}0.302 & \cellcolor{yellow!40}0.184 \\
posner.a.fwd & 255 & 2.090 & 9.258 & 2.289 & \cellcolor{gray!20}1.930 & \cellcolor{yellow!40}1.714 & 0.774 & 1.500 & 0.714 & \cellcolor{gray!20}0.653 & \cellcolor{yellow!40}0.584 \\

\midrule
cfa.1.lat & 82 & 1.452 & 1.508 & 1.724 & \cellcolor{yellow!40}0.418 & \cellcolor{gray!20}0.717 & 0.780 & 1.101 & 2.122 & \cellcolor{yellow!40}0.298 & \cellcolor{gray!20}0.486 \\
cfa.3.lat & 67 & 0.775 & 0.987 & 1.794 & \cellcolor{yellow!40}0.401 & \cellcolor{gray!20}0.552 & 0.794 & 1.236 & 15.175 & \cellcolor{yellow!40}0.370 & \cellcolor{gray!20}0.372 \\
cfa.a.lat (1) & 76 & 0.984 & 3.230 & 1.022 & \cellcolor{gray!20}0.372 & \cellcolor{yellow!40}0.255 & 0.710 & 1.647 & 1.548 & \cellcolor{gray!20}0.386 & \cellcolor{yellow!40}0.355 \\
cfa.a.lat (2) & 77 & 0.538 & 1.076 & 2.479 & \cellcolor{gray!20}0.259 & \cellcolor{yellow!40}0.161 & 0.642 & 0.986 & 2.577 & \cellcolor{gray!20}0.387 & \cellcolor{yellow!40}0.233 \\
morrison.1.lat (1) & 52 & 0.831 & 0.844 & 2.004 & \cellcolor{yellow!40}0.208 & \cellcolor{gray!20}0.355 & 0.739 & 0.726 & 1.954 & \cellcolor{yellow!40}0.312 & \cellcolor{gray!20}0.350 \\
morrison.1.lat (2) & 106 & 1.326 & 4.087 & 9.302 & \cellcolor{yellow!40}0.563 & \cellcolor{gray!20}0.751 & 0.975 & 1.649 & 6.239 & \cellcolor{yellow!40}0.616 & \cellcolor{gray!20}0.662 \\
morrison.2.lat & 75 & 1.055 & 1.385 & 3.601 & \cellcolor{gray!20}0.854 & \cellcolor{yellow!40}0.688 & 0.860 & 0.987 & 3.563 & \cellcolor{gray!20}0.452 & \cellcolor{yellow!40}0.439 \\
posner.1.lat & 338 & 6.807 & 34.347 & 3.919 & \cellcolor{gray!20}2.185 & \cellcolor{yellow!40}0.885 & 0.764 & 1.510 & 2.154 & \cellcolor{gray!20}0.507 & \cellcolor{yellow!40}0.391 \\
posner.2.lat & 207 & 1.899 & 2.857 & \cellcolor{yellow!40}1.195 & \cellcolor{gray!20}1.293 & 1.709 & 0.616 & 1.322 & 1.391 & \cellcolor{yellow!40}0.414 & \cellcolor{gray!20}0.422 \\
posner.3.lat & 205 & 1.541 & 1.468 & 1.492 & \cellcolor{yellow!40}1.085 & \cellcolor{gray!20}1.112 & 0.642 & 1.400 & 1.247 & \cellcolor{gray!20}0.438 & \cellcolor{yellow!40}0.438 \\
posner.a.lat & 248 & 4.714 & 8.113 & \cellcolor{yellow!40}4.091 & 5.810 & \cellcolor{gray!20}4.187 & 0.948 & 1.426 & \cellcolor{gray!20}0.784 & 0.839 & \cellcolor{yellow!40}0.724 \\
\bottomrule

\end{tabular}
}
\label{tab:results_fwd}

    \vspace{-5mm}
\end{table*}

\subsection{Pose Graph Optimization}
\label{sec:pgo}
To fuse our predicted radar measurements with IMU for state estimation, we follow MRIO\cite{huang2024multi}. Trajectory estimation is performed by a sliding-window factor graph over a 3-second window. At each radar frame, the network produces a body-frame velocity measurement $\hat{\vv} | \hat{\vv}_d$ and covariance $\Sigma_v|\Sigma_d$ from~\eqref{eq:wls}--\eqref{eq:cov}. These are incorporated as velocity factors replacing the radar Doppler point cloud estimated velocity and uncertainty used in RIO systems. 
\begin{equation}
    \begin{aligned}
    \Xc^*_\Sc &= \argmin_{\Xc_\Sc}\Bigg[ \sum_{i,j\in\Sc}\Big( ||\rv_{\Delta\Iv_{ij}}||^2_{\Sigma_\rv} + ||\rv_{\Vv^b_{j}}||^2_{\Sigma_{\vv^b}}\Big)\Bigg]
    \end{aligned}
\end{equation}
To optimize poses $\Xc\in\Rb^6$ in a sliding window $\Sc$ between frame $i$ and frame $j$, we minimize the IMU residuals $\rv_{\Delta\Iv_{ij}}$, where IMU measurements between consecutive radar frames are integrated via preintegration to form relative-motion factors~\cite{forster2017imu}, and the body-frame velocity residuals $\rv_{\Vv^b_{j}}$ following~\cite{huang2024multi}. We use the estimated covariance from the neural network to minimize the Mahalanobis distance $||\dots||^2_{\Sigma_{\vv^b}}$. A Huber robust cost is applied to the velocity residuals to limit the influence of outlier measurements. States outside the optimization window are removed via marginalization with a fixed prior on all six pose degrees of freedom. The system runs online without any loop closure or global map.

\begin{figure*}[t]
    \centering
    \includegraphics[width=\textwidth]{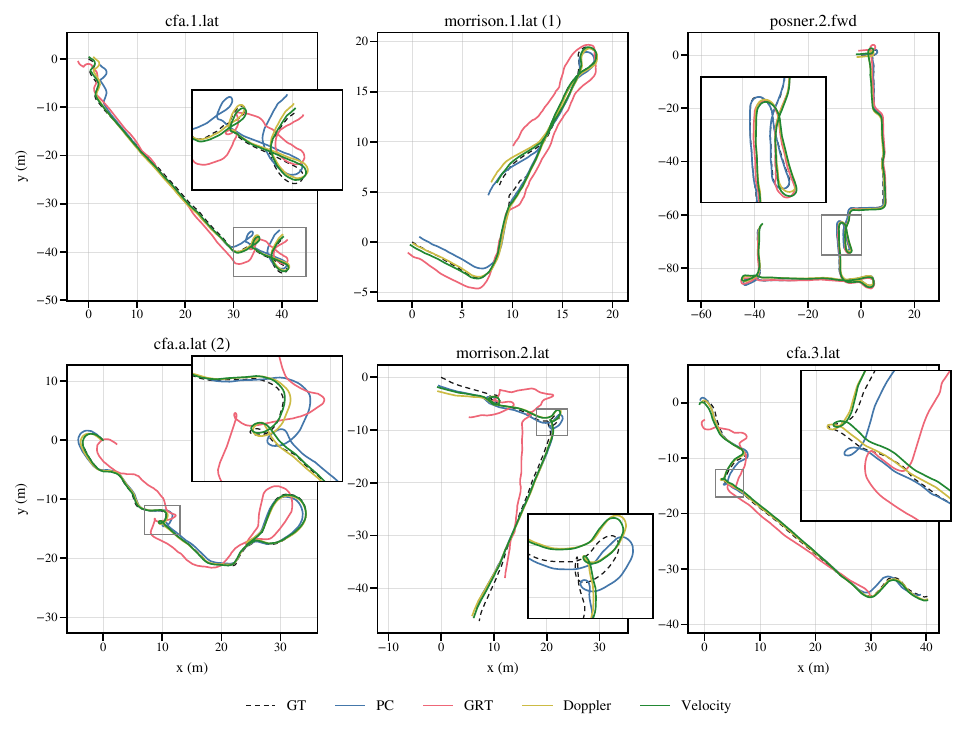}
    \vspace{-10mm}
    \caption{\textbf{Qualitative trajectory comparison on six representative  sequences.}  Ground truth is shown as a dashed black line; estimated trajectories from PC, GRT, Doppler, and Velocity are shown as solid colored lines, aligned to the ground truth via SE(3) Umeyama alignment. Insets highlight regions of interest where method differences are most pronounced. Lateral-motion sequences demonstrate the clear advantage of learning-based methods over the classical DSP point-cloud baseline.}
    \label{fig:qual}
    \vspace{-5mm}
\end{figure*}

\section{Experiments}
\label{sec:exp}
\subsection{Radar IQ Dataset}
We train and evaluate our proposed methods with the IQ1M dataset~\cite{huang2025towards}, which contains about 29 hours, 1 million frames at 20 frames per second of complex signal data collected from a Texas Instruments AWR1843Boost radar and a DCA-1000EVM capture card. The IMU measurements are provided by an XSens MTi-3 AHRS sensor running at 100 frames per second. The ground truth depth and velocity are provided by the dataset which were built using the Cartographer~\cite{hess2016real} LiDAR state estimation system. To demonstrate generalizability to a different chirp configuration, platform, and setting, we also fine-tune and evaluate on ColoRadar~\cite{kramer2022coloradar} (Sec.~\ref{sec:coloradar}), which adds scenarios of outdoor environments and velocity profiles beyond its unambiguous Doppler interval.


\ieeepar{Evaluation Subset.} The IQ1M dataset covers three scene and motion regimes (indoor, outdoor, bike), each with a different chirp configuration. As described in Sec.~\ref{sec:nn_train}, we pretrain on the full training set. The velocity and uncertainty stages are then trained on the indoor subset alone, which is most relevant to state-estimation since it is the most cluttered and varied, and the only subset with both forward and lateral traces.


\ieeepar{2D Evaluation.} The AWR1843Boost antenna is designed for automotive applications, where the scene is planar; as such, its antenna geometry heavily favors azimuth in angular resolution (8 azimuth vs 2 elevation bins) and field of view ($\pm 60^\circ$ -3dB-beamwidth horizontally vs $\pm 30^\circ$ vertically). This causes the elevation to be extremely noisy; as such, we limit estimation to the horizontal plane by fixing the marginalized pose to $z=0$.

\subsection{Baselines}
We compare against EKF-RIO~\cite{DoerMFI2020}, a well-established radar-inertial odometry system, and two baseline velocity estimators that we run inside our own odometry system, changing only the front end.

\ieeepar{CFAR point cloud.}
\label{sec:radar_dsp}
Following a commonly-used classical radar signal processing pipeline \cite{iovescu2017fundamentals}, we convert raw IQ samples into a 3-D point cloud through three sequential stages\footnote{
All steps are implemented in PyTorch with GPU acceleration.
}: a 4-D FFT\footnote{
Note that we also apply a DC-removal step by subtracting the mean along the slow-time dimension before the FFT, which removes zero-Doppler clutter caused by antenna and device artifacts.
} as described in section~\ref{sec:signal_processing}, 2-D cell-averaging CFAR to detect candidate objects in the Range-Doppler map, and angle of arrival estimation by selecting the peak-magnitude bin $\psi_n$ in its corresponding elevation-azimuth FFT, which we map to a physical angle via
\begin{equation}
  \theta_n = \arcsin\!\left(\frac{\psi_n}{2\pi d_\lambda}\right)
\end{equation}                              
where $d_\lambda$ is the antenna spacing in wavelengths\footnote{
This is typically $0.5$ by design, but may vary with the operating frequency.
}.

Using the (range, Doppler, azimuth, elevation) information for each detection, we then transform them into 4-D point clouds $(x, y, z, \text{Doppler})$. Finally, to suppress noise and multipath detections, we discard points outside the $-3\,\mathrm{dB}$ radiation-pattern lobe, which spans approximately $(\pm60^\circ\ \text{azimuth},\ \pm30^\circ\ \text{elevation})$ for the AWR1843Boost.

Following~\cite{huang2024multi}, the linear body-frame velocity and its covariance are then computed from the resulting point cloud and subsequently fused with IMU measurements to estimate pose. This approach marked as PC in later performance comparisons.

\ieeepar{EKF-RIO.}
Using the CFAR point cloud, we also run EKF-RIO~\cite{DoerMFI2020}, an error-state extended Kalman filter that fuses IMU strapdown integration with 3-DoF radar ego-velocity measurements. Each radar scan is reduced to a single body-frame velocity by the RANSAC/LSQ estimator. We build the authors' released implementation against a ROS-free compatibility layer so it consumes our CFAR detections, and we drive it with the same IMU noise model. 

\ieeepar{GRT output velocity.}
Using a GRT model~\cite{huang2025towards} pretrained for velocity prediction, we convert its' Doppler resolution-normalized velocity back to metric units, and apply a fixed covariance of $\Sigma_v = \mathrm{diag}(0.01, 0.01, 0.01) \in \Rb^{3\times3}$ before applying our odometry system (Sec.~\ref{sec:pgo}).

\subsection{Quantitative Evaluation}
We evaluate our radar-inertial odometry using the EVO~\cite{grupp2017evo} library, and measure Absolute Pose Error (APE) computed with Umeyama alignment and Relative Pose Error (RPE) evaluated at 10-meter intervals.

UNRIO achieves the lowest RPE in the majority of sequences (Table~\ref{tab:results_fwd}), demonstrating superior local accuracy compared to all baselines. While UNRIO outperforms baselines on the majority of forward-motion sequencies, its advantage is particularly evident in the lateral-motion sequences, where learning-based estimators consistently and substantially outperform the point cloud baseline. While UNRIO approach leverages the full Doppler spectrum to maintain accuracy, classical point cloud RSP produces fewer points along the direction of motion when the sensor is moving laterally. Since the body velocity must be fit from few detections in a narrow azimuth band, and the lateral velocity component projects weakly onto the \textit{radial} velocity measured by Doppler, the fit is ill-conditioned, leading to higher noise and poor performance.

\subsection{Qualitative Evaluation}
UNRIO's advantages over point-cloud baselines can also be clearly seen qualitatively (Fig.~\ref{fig:qual}),  as in the lateral sequences, \texttt{PC} and \texttt{EKF-RIO} point-cloud baselines drift heavily from the ground truth. On the other hand, the \texttt{GRT} baseline is competitive on some traces, yet fails on others. Its predictions are less accurate due to its model design and training, while its fixed covariance leaves the back end no way to discount noisy measurements. In contrast, \texttt{UNRIO-Doppler} and \texttt{UNRIO-Velocity} track the ground truth throughout, with \texttt{UNRIO-Velocity} holding tightest where the others diverge as shown by the insets.

\subsection{Generalization to a Different Chirp Configuration}
\label{sec:coloradar}
We further evaluate on the ColoRadar~\cite{kramer2022coloradar} dataset, which provides raw ADC samples\footnote{
Since ColoRadar includes detections generated by TI's hardware RSP implementation, we use these rather than our own CFAR pipeline for point cloud baselines. 
} from a TI AWR1843 single-chip radar together with an IMU\footnote{
Due to unstable IMU initialization, we discard the first 10s of each sequence during evaluation.
}, a LiDAR, and ground-truth poses, recorded on a handheld platform across indoor laboratory, hallway, classroom, and outdoor scenes. Using a $80\%/20\%$ train/test split holding out the last few runs of each scene, we fine-tune both UNRIO models (Sec.~\ref{sec:vel_pred}) on the ColoRadar training split, leaving the architecture and the odometry pipeline unchanged.

Although the radar chip is the same as in IQ1M, its waveform configuration differs substantially, with 128 range bins over 16m compared to 256 bins over 11.2m in IQ1M indoor subset, 128 Doppler bins over 2.5m/s compared to 64 bins over 1.2m/s in IQ1M-indoor, and a frame rate at 10 Hz versus 20 Hz in IQ1M. By fine-tuning on the relatively limited training split, UNRIO is nevertheless able to achieve better RPE (Table~\ref{tab:results_all}) than baselines.

\begin{figure*}[t]
    \centering
    \includegraphics[width=\textwidth]{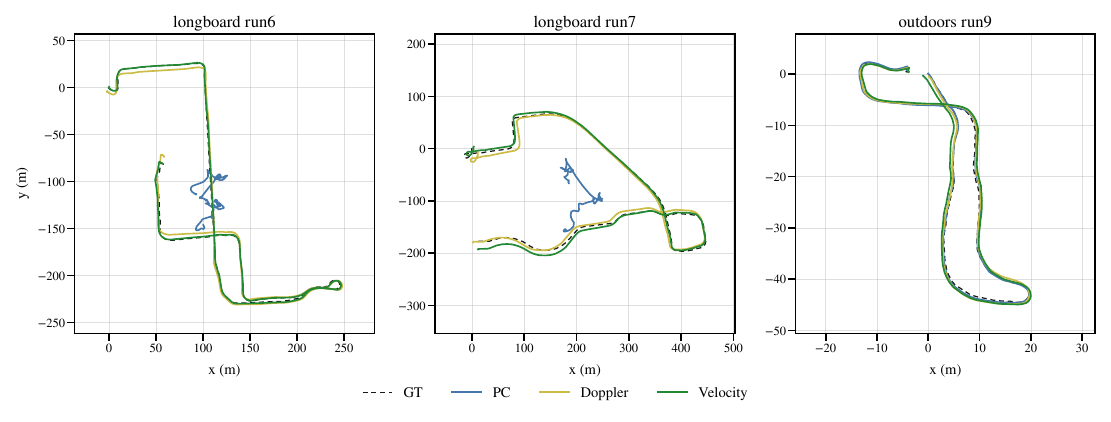}
    \vspace{-10mm}
    \caption{%
    \textbf{ColoRadar trajectories vs.\ ground truth (dashed)} The radar is Doppler-unambiguous to $\pm2.50$\,m/s; the longboard rides run at a median $4.7$ and $5.5$\,m/s, aliasing $80\%$ of frames. The point-cloud baseline reads Doppler directly and underestimate the velocity, while both learned models stay bounded.}
    \label{fig:colo_traj}
\end{figure*}

The two longboard test sequences also move far faster than the waveform can unambiguously measure: their mean speed is $4.3$ and $5.2$~m/s with peaks above $8$~m/s, and $\approx80\%$ of their frames fall outside the $\pm2.50$~m/s unambiguous Doppler limit and alias in the spectrum. This leads to substantially degraded performance for point-cloud baselines, which manifests in state estimation failure (Fig.~\ref{fig:colo_traj}), while UNRIO remains robust to Doppler aliasing.

\begin{table*}[t]
\centering
\small
\caption{APE / RPE RMSE (m). --- ColoRadar test sequences evaluation. Color background indicate \colorbox{yellow!40}{\strut best}, \colorbox{gray!20}{\strut 2nd best} performance.}
\vspace{-2mm}
\resizebox{0.95\textwidth}{!}{%
\begin{tabular}{lrrrrr|rrrr}
\toprule
Trace & Dist (m) & \multicolumn{4}{c}{APE} & \multicolumn{4}{c}{RPE} \\
\cmidrule{3-6} \cmidrule{7-10}
 &  & PC~\cite{huang2024multi} & \makecell{EKF\\RIO}~\cite{DoerMFI2020} & \makecell{UNRIO\\Doppler} & \makecell{UNRIO\\velocity} & PC~\cite{huang2024multi} & \makecell{EKF\\RIO}~\cite{DoerMFI2020} & \makecell{UNRIO\\Doppler} & \makecell{UNRIO\\velocity} \\
\midrule
12\_21\_2020\_arpg\_lab\_run4 & 78 & \cellcolor{yellow!40}0.400 & \cellcolor{gray!20}0.468 & 1.368 & 0.780 & 0.908 & \cellcolor{gray!20}0.871 & 1.400 & \cellcolor{yellow!40}0.809 \\
12\_21\_2020\_ec\_hallways\_run4 & 93 & \cellcolor{yellow!40}0.464 & 0.978 & 1.038 & \cellcolor{gray!20}0.867 & 1.234 & 1.349 & \cellcolor{gray!20}0.449 & \cellcolor{yellow!40}0.345 \\
2\_22\_2021\_longboard\_run6 & 846 & 102.078 & 528.885 & \cellcolor{gray!20}4.835 & \cellcolor{yellow!40}1.260 & 36.308 & 11.912 & \cellcolor{gray!20}0.525 & \cellcolor{yellow!40}0.377 \\
2\_22\_2021\_longboard\_run7 & 1201 & 146.366 & 6181.662 & \cellcolor{gray!20}9.317 & \cellcolor{yellow!40}6.452 & 33.939 & 11.321 & \cellcolor{gray!20}0.754 & \cellcolor{yellow!40}0.656 \\
2\_23\_2021\_edgar\_army\_run5 & 103 & \cellcolor{gray!20}0.399 & 1.977 & 0.515 & \cellcolor{yellow!40}0.338 & 1.211 & 1.919 & \cellcolor{gray!20}0.501 & \cellcolor{yellow!40}0.336 \\
2\_23\_2021\_edgar\_classroom\_run5 & 181 & \cellcolor{gray!20}1.735 & \cellcolor{yellow!40}1.700 & 2.816 & 2.198 & 1.134 & 1.919 & \cellcolor{gray!20}0.469 & \cellcolor{yellow!40}0.412 \\
2\_24\_2021\_aspen\_run10 & 82 & \cellcolor{yellow!40}0.321 & 0.453 & 0.542 & \cellcolor{gray!20}0.334 & 0.577 & 0.704 & \cellcolor{gray!20}0.538 & \cellcolor{yellow!40}0.302 \\
2\_24\_2021\_aspen\_run11 & 79 & \cellcolor{gray!20}0.512 & 0.532 & 0.525 & \cellcolor{yellow!40}0.404 & 0.753 & 1.089 & \cellcolor{gray!20}0.610 & \cellcolor{yellow!40}0.569 \\
2\_28\_2021\_outdoors\_run8 & 112 & 1.891 & 2.072 & \cellcolor{gray!20}0.563 & \cellcolor{yellow!40}0.351 & 1.949 & 2.901 & \cellcolor{gray!20}0.369 & \cellcolor{yellow!40}0.269 \\
2\_28\_2021\_outdoors\_run9 & 140 & \cellcolor{yellow!40}0.248 & 1.342 & \cellcolor{gray!20}0.425 & 0.581 & 1.349 & 1.659 & \cellcolor{yellow!40}0.399 & \cellcolor{gray!20}0.419 \\
\bottomrule
\end{tabular}
}
\label{tab:results_all}
\vspace{-3mm}
\end{table*}

\subsection{Ablation Studies on Uncertainty Prediction}
\label{sec:ablation}

To test whether the predicted uncertainty helps, we replace the per-frame covariance in the radar-velocity factor with a constant fitted on the training split, using one trace per distinctive scene:\footnote{Fitting instead on several traces of a single scene yields a constant tuned to that scene's dynamics; one trace per scene keeps the fast longboard rides represented in the ColoRadar constant.} $\bar\Sigma_\text{net}$, the mean of the model's own predicted covariance, and $\bar\Sigma_\text{emp}$, the empirical covariance of the velocity residual, which a practitioner would tune by hand. The Doppler branch uses the learned uncertainty a second time, as the per-pixel weights $W$ of the least-squares fit, which we ablate on its own and jointly with the covariance.

Comparing the sequences won by each method as measured by RPE (10m), we find that the two branches behave oppositely (Table~\ref{tab:ablation}). The velocity branch regresses its log-variance under a Gaussian likelihood in $(\text{m/s})^2$, and its per-frame covariance wins 22 of 23 IQ1M sequences against $\bar\Sigma_\text{net}$ and 18 of 23 against $\bar\Sigma_\text{emp}$; ColoRadar's ten sequences do not separate the arms. The Doppler branch reuses the per-pixel depth log-variance, supervised on a depth residual in meters and therefore mis-scaled as a velocity uncertainty; its per-frame covariance never wins, tying both constants on IQ1M and losing 9 of 10 on ColoRadar.

The depth log-variance also fails as a weighting. With the covariance held at $\bar\Sigma_\text{emp}$ on both arms so that only $\hat{v}_d$ varies, it beats uniform weights on 5 of 23 IQ1M sequences; this tests the confidence \emph{ordering} alone, independent of the scale mismatch.\footnote{$\hat{v}_d = (D^\top W D)^{-1} D^\top W(-d)$ is unchanged by $W \rightarrow \alpha W$, so this arm is insensitive to the scale of the predicted uncertainty.} Ablating both uses leaves no learned uncertainty anywhere -- uniform weights and $\Sigma = (D^\top D)^{-1}$\footnote{Under uniform weights the fit reduces to ordinary least squares and $(D^\top D)^{-1}$ is one constant matrix for the whole dataset, set by the angular field of view of the direction grid. Its diagonal, $\mathrm{diag}(7.7, 17.9, 53.2) \times 10^{-4}$, is anisotropic only because the elevation field of view is half the azimuth one.} -- and the full pipeline still wins only 7 of 23. We attribute this to \emph{depth} log-variance being uninformative in precisely the regimes that corrupt ego-velocity: moving scatterers and Doppler aliasing both yield confident depth with an unreliable radial velocity, which suggests supervising the Doppler log-variance channel directly as the natural remedy. What stands in the way is the label: the per-pixel Doppler target is a rigid-body projection of the ground-truth body velocity, so it is correct only under a static world and is wrong on exactly the dynamic pixels whose uncertainty we would want to learn. Training such a map therefore requires a label carrying the true Doppler velocity of dynamic objects, which we leave to future work.

\begin{table}[t]
\caption{Uncertainty ablation: sequences won on RPE (10\,m).}
\label{tab:ablation}
\centering
\resizebox{\columnwidth}{!}{%
\begin{tabular}{llcc}
\toprule
UNRIO & Ablated & ColoRadar & IQ1M \\
\midrule
\multicolumn{4}{l}{\textit{Covariance, velocity branch}} \\
per-frame $\Sigma$ & $\bar\Sigma_\text{net}$ & 5 / 5 & \textbf{22 / 1} \\
per-frame $\Sigma$ & $\bar\Sigma_\text{emp}$ & 7 / 3 & \textbf{18 / 5} \\
\midrule
\multicolumn{4}{l}{\textit{Covariance, Doppler branch}} \\
per-frame $\Sigma$ & $\bar\Sigma_\text{net}$ & 1 / 9 & 10 / 13 \\
per-frame $\Sigma$ & $\bar\Sigma_\text{emp}$ & 1 / 9 & 14 / 9 \\
\midrule
\multicolumn{4}{l}{\textit{Least-squares weighting, Doppler branch}} \\
depth $w$, $\bar\Sigma_\text{emp}$ & uniform $w$, $\bar\Sigma_\text{emp}$ & 4 / 6 & 5 / 18 \\
depth $w$, per-frame $\Sigma$ & uniform $w$, $(D^\top D)^{-1}$ & 4 / 6 & 7 / 16 \\
\bottomrule
\end{tabular}%
}
\end{table}

\subsection{Runtime Analysis}

The full UNRIO system runs fully real-time, with the FFT front-end, neural network inference, and pose-graph optimization backend able to run in ${\sim}25\,\mathrm{ms}$ in fp32, and ${\sim}15\,\mathrm{ms}$ (about $1.6\times$ faster) in bf16. Crucially, both remain well within the radar frame period  ($50$--$100\,\mathrm{ms}$).

\begin{table}[t]
  \centering
  \caption{End-to-end pipeline (FFT + network + PGO) runtime per frame in ms (RTX~4090, Intel
  i9-13900K, batch~1).}
  \label{tab:pipeline-runtime}
  \begin{tabular}{lllrrrr}
    \toprule
    & & & \multicolumn{2}{c}{fp32} & \multicolumn{2}{c}{bf16} \\
    \cmidrule(lr){4-5}\cmidrule(lr){6-7}
    Dataset & FFT & Stage & Doppler & Vel & Doppler & Vel \\
    \midrule
    ColoRadar & 8.98 & Network        & 15.85 & 15.67 &  5.78 &  5.58 \\
              &      & \textbf{Total} & 25.41 & 25.22 & 15.34 & 15.14 \\
    \midrule
    IQ1M      & 9.19 & Network        & 15.89 & 15.69 &  6.17 &  5.59 \\
              &      & \textbf{Total} & 25.65 & 25.46 & 15.94 & 15.36 \\
    \bottomrule
  \end{tabular}
\end{table}

\section{Conclusion}
We presented UNRIO, a radar-inertial odometry system that estimates ego-velocity and its uncertainty directly from raw mmWave radar IQ signals, bypassing the information loss inherent in classical signal processing pipelines. A transformer-based network predicts velocity either directly or as a per-angle-bin Doppler map, and its uncertainty is propagated into a sliding-window pose graph fusing radar and IMU measurements. Evaluated on the public benchmarks IQ1M and ColoRadar, UNRIO achieves the lowest relative pose error across held-out environments spanning different chirp configurations, motion patterns, and platforms. Its advantage is most pronounced where point-cloud methods break down — lateral motion in IQ1M and severe Doppler aliasing in ColoRadar. Our ablation suggests the learned depth uncertainty reused for the Doppler fit is not an accurate velocity uncertainty, but an explainable heuristic, and that predicting velocity and its uncertainty directly does better here. Operating on the full spectrum nevertheless recovers accurate velocity in both.

\ieeepar{Limitations}. The Doppler branch still lacks a dedicated uncertainty head of its own. IQ1M and ColoRadar also share the same radar sensor despite differing chirp configurations, leaving generalization across antenna layouts untested. Finally, the lack of elevation resolution and field of view in available datasets has limited evaluation of UNRIO to 2D. We leave these limitations to future work.

\bibliographystyle{IEEEtran}
\balance
\bibliography{ref}

\end{document}